\documentclass[runningheads]{llncs}
\usepackage{bbding}

\usepackage{marvosym}
\usepackage{float}
\usepackage{amsmath}
\usepackage{dutchcal}
\usepackage{hyperref}
\usepackage{subfigure}
\usepackage{graphicx}
\usepackage{color}
\usepackage{amssymb}
\usepackage{booktabs}
\usepackage{makecell}
\usepackage{svg}
\usepackage{datetime}
\usepackage{multirow}
\usepackage{cases}
\usepackage{textcomp}

\newcommand{\figref}[1]{Fig.~\ref{#1}}

\begin{document}
\title{Multi-modal Instance Refinement for
Cross-domain Action Recognition}

\author{Yuan Qing\inst{1} \and
Naixing Wu\inst{2} \and
Shaohua Wan\inst{1} \and Lixin Duan\inst{1}\textsuperscript{(\Letter)}}
\vspace{-5pt}
\authorrunning{Y. Qing et al.}
%
\institute{Shenzhen Institute for Advanced Study, University of Electronic Science and Technology of China \\ \and China Unicom Shenzhen Branch \\ \email{yuanqing.shawn@gmail.com}, \email{1320544942@qq.com}, \email{\{shaohua.wan, lxduan\}@uestc.edu.cn}}

%
%

%

%


%
\maketitle              

\vspace{-16pt}
\begin{abstract}
Unsupervised cross-domain action recognition aims at adapting the model trained on an existing labeled source domain to a new unlabeled target domain. Most existing methods solve the task by directly aligning the feature distributions of source and target domains. However, this would cause negative transfer during domain adaptation due to some negative training samples in both domains. In the \textit{source domain}, some training samples are of low-relevance to target domain due to the difference in viewpoints, action styles, etc. In the \textit{target domain}, there are some ambiguous training samples that can be easily classified as another type of action under the case of source domain. The problem of negative transfer has been explored in cross-domain object detection, while it remains under-explored in cross-domain action recognition. Therefore, we propose a \textit{Multi-modal Instance Refinement (MMIR)} method to alleviate the negative transfer based on reinforcement learning. Specifically, a reinforcement learning agent is trained in both domains for every modality to refine the training data by selecting out negative samples from each domain. Our method finally outperforms several other state-of-the-art baselines in cross-domain action recognition on the benchmark EPIC-Kitchens~\cite{epic} dataset, which demonstrates the advantage of MMIR in reducing negative transfer. 
\vspace{-5pt}
\end{abstract}
\vspace{-20pt}
\section{Introduction}
\begin{figure}[t]
\vspace{-3pt}
\centering

\subfigure[Less-relevant action]{
\label{comparea}
\centering

\rotatebox{90}{\qquad \, \footnotesize{Target} \qquad \qquad \; \footnotesize{Source}}

\begin{minipage}[b]{.36\linewidth}

\centering

\includegraphics[width=1.8in]{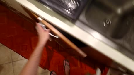} \\
\includegraphics[width=1.8in]{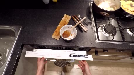}

\end{minipage}%
}
\subfigure[Ambiguous action]{
\label{compareb}
\begin{minipage}[b]{.36\linewidth}
\centering
\includegraphics[width=1.8in]{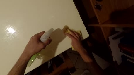} \\
\includegraphics[width=1.8in]{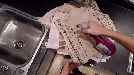}
\end{minipage}%
}
\vspace{-12pt} 
\caption{Negative samples that cause negative transfer from source domain \textbf{D2} (first row) and target domain \textbf{D3} (second row) defined in~\cite{mmsada}.}

\label{compare}
\vspace{-20pt}
\end{figure}

Action recognition is one of the most important tasks in video understanding, which aims at recognizing and predicting human actions in videos~\cite{internvideo,mvd,i3d,epic}. The majority of the works in action recognition are carried out on the basis of supervised learning, which involves using a large number of labeled video/action segments to train a model for a specified scene setting. However, obtaining a large number of annotated video data for a certain scene would be very costly and sometimes difficult due to the environment setup and video post-processing as well as labeling. To fully leverage existing labeled data and reduce the cost of acquiring new data, Unsupervised Domain Adaptation (UDA)~\cite{mcd,uda2,udalong2} has been introduced to generalize a model trained on a source domain with adequate annotations to a target domain with no labels, where the two domains differentiate from each other but are partially related.  
For action recognition, though there are several UDA methods~\cite{sourcefree,constra,mmsada} proposed, most of them achieve this by directly aligning the feature distribution of source and target domains. However, this could lead to negative transfer during domain adaptation due to some negative training samples in both domains~\cite{sir}. For instance, there might be some ambiguous actions in target domain that do not belong to the defined action categories in the source domain or are very similar to another kind of action in the source domain. Additionally, there might also be some less-relevant actions in source domain that have completely different viewpoints/action styles compared with samples in the target domain. To be specific, \figref{compare} shows these two types of negative samples in domain D2 (source) and D3 (target) defined in~\cite{mmsada} from EPIC-Kitchens dataset~\cite{epic}. In \figref{comparea}, action \emph{open} in source domain is considered less-relevant to that of the target domain since the trajectory of motion and way of opening are dissimilar. In \figref{compareb}, a spraying action in target domain that does not belong to a predefined action type is likely to be mistakenly recognized as \emph{wipe} due to the similarity in action style and appearance. 

To alleviate the impact of negative transfer brought by these negative training samples, we propose Multi-modal Instance Refinement (MMIR) based on deep Q-learning (DQN)~\cite{DQN}, under the framework of MM-SADA~\cite{mmsada}. Our MMIR trains reinforcement learning agents in both domains in each modality to refine source and target samples by selecting out less-relevant source instances from source domain and ambiguous target instances from target domain. To the best of our knowledge, there's no previous work on reducing negative transfer in cross-domain action recognition. Our contributions are summarised as follows:
\begin{itemize}
\vspace{-6pt}
    \item As far as we know, we are the first to define and tackle the issue of negative transfer in cross-domain action recognition.
    \item We adopt a novel instance refinement strategy using deep reinforcement learning to select outlier instances in source domain and target domain within two modalities (RGB and optical flow). 
    \item Our method achieves superior performance compared with other state-of-the-art methods in cross-domain action recognition on EPIC-Kitchens dataset.
\end{itemize}

\section{Related Work}
\subsection{Action Recognition}

In action recognition, early works use 2D/3D convolution~\cite{early2dact,early3dact} for feature extraction only in a single modality, i.e., RGB. Later, optical flow of video segments is used as auxiliary training data which carries more temporal and motion information compared with RGB~\cite{optical}. Therefore, current popular CNN-based methods adopt a two-stream 3D convolutional neural network structure~\cite{two3d,i3d} for feature extraction which could utilize the information contained in multiple modalities and model the temporal information. Most recently, vision transformer~\cite{vit} based approaches have excelled CNN-based methods on many benchmarks. MVD~\cite{mvd} builds a two-stage masked feature modeling framework, where the high-level features of pretrained models learned in the first stage will serve as masked prediction targets for student model in the second stage. Although these methods show promising performance in a supervised manner, we are going to focus on action recognition under the setting of UDA.

\vspace{-10pt}
\subsection{Unsupervised Domain Adaptation for Action Recognition}
Though both RGB and optical flow have been studied for domain adaptation in action recognition, there are only a limited number of works attempted to conduct multi-modal domain adaptation~\cite{sourcefree,constra,mmsada}. Munro and Dame~\cite{mmsada} propose MM-SADA, a multi-modal 3D convolutional neural network with a self-supervision classifier between modalities. It uses Gradient Reversal Layer (GRL)~\cite{uda} to implement domain discriminator within different modalities. Kim \emph{et al.}~\cite{constra} apply contrastive learning to design a unified framework using transformer for multi-modal UDA in video understanding. Xu \emph{et al.}~\cite{sourcefree} propose a source-free UDA model to learn temporal consistency in videos between source domain and target domain. Similar to~\cite{mmsada}, our work adopts a multi-modal 3D ConvNet for feature extraction and utilizes domain adversarial learning, but we focus on a different task by incorporating deep reinforcement learning into our action recognition framework to eliminate the effect of negative transfer.

\vspace{-10pt}
\subsection{Deep Reinforcement Learning}
Deep reinforcement learning has been applied to various tasks in computer vision~\cite{rl1,rl3}. Wang \emph{et al.}~\cite{rl1} design a reinforcement learning-based two-level framework for video captioning, in which a low-level module recognizes the original actions to fulfill the goals specified by the high-level module. ReinforceNet~\cite{rl3} incorporates region selection and bounding box refinement networks to form a reinforcement learning framework based on CNN to select optimal proposals and refine bounding box positions. Recently, several works apply reinforcement learning to action recognition~\cite{rlact2,rlact3}. Dong \emph{et al.}~\cite{rlact2} design a deep reinforcement learning framework to capture the most discriminative frames and delete confusing frames in action segments. Weng \emph{et al.}~\cite{rlact3} improve recognition accuracy by designing agents that learn to produce binary masks to select out interfering categories. All these methods adopt deep reinforcement learning to refine negative frames in the action segment within the same domain, while our method uses deep reinforcement learning to refine negative action segments across domains to handle negative transfer in cross-domain action recognition.

\section{Proposed Method}

In unsupervised cross-domain action recognition, two domains are given, namely source and target. A labeled source domain  $\mathcal{D}^s$ is denoted as $\mathcal{D}^s = \{(\mathcal{x}^s_i, \mathcal{y}^s_i)|^{\mathcal{N}_s}_{i=1}\}$, where $\mathcal{x}^s_i$ is the $\mathcal{i}$-th action segment and $\mathcal{y}^s_i$ is its verbal label. Similarly, the unlabeled target domain $\mathcal{D}^t$ is represented by $\mathcal{D}^t = \{\mathcal{x}^t_i|^{\mathcal{N}_t}_{i=1}\}$, where $\mathcal{x}^t_i$ is the $\mathcal{i}$-th action segment. For action segments, each segment is formed as a sequence of $L$ frames. Therefore, we have $\mathcal{x}^s_i = \{ \mathcal{x}_{i,1}^s,\mathcal{x}_{i,2}^s,\mathcal{x}_{i,3}^s,\ldots,\mathcal{x}_{i,L}^s \}$ and $\mathcal{x}^t_i = \{ \mathcal{x}_{i,1}^t,\mathcal{x}_{i,2}^t,\mathcal{x}_{i,3}^t,\ldots,\mathcal{x}_{i,L}^t \}$, respectively.

To reduce negative transfer during domain adaptation, two reinforcement learning agents, S-agent and T-agent, are defined under a deep-Q learning network (DQN) to make selections in source and target domains. S-agent learns policies to select out less-relevant action segments from source action segments $\mathcal{x}^s$, while T-agent is trained to select out ambiguous action segments from target action segments $\mathcal{x}^t$. After refinement, we use the refined instances $\hat{\mathcal{x}^s}$ and $\hat{\mathcal{x}^t}$ to train our domain discriminator to learn domain invariant features.

\begin{figure}[!ht]

    \centering
    \subfigure[Overall model structure]{
    \begin{minipage}[ht]{1\linewidth}
    \centering
    \includegraphics[width=3.5in]{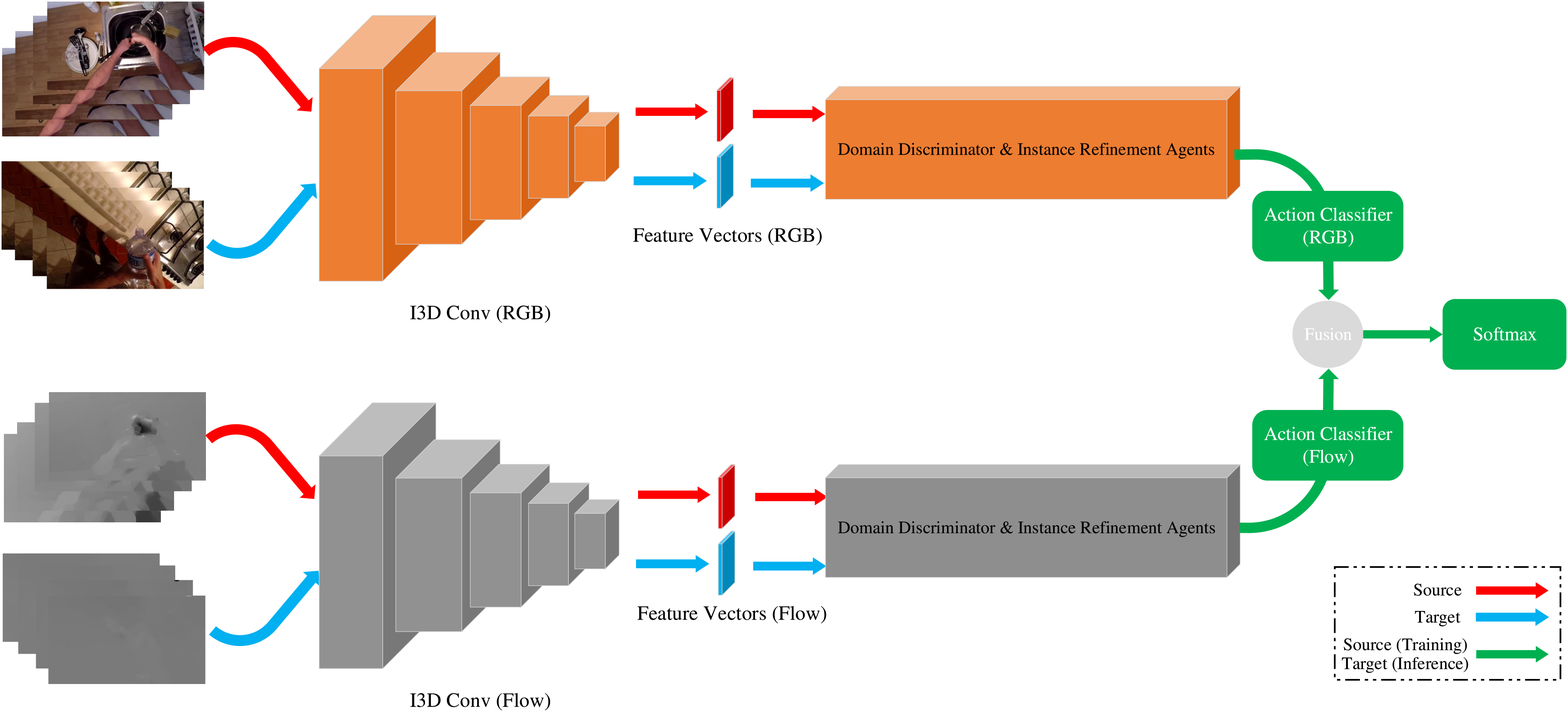}
    \vspace{5pt}
    \label{overall}
    \end{minipage}
        }
    
    \subfigure[Domain discriminator and instance refinement agents]{
    \begin{minipage}[ht]{1\linewidth}
    \centering
        \includegraphics[width=3.5in]{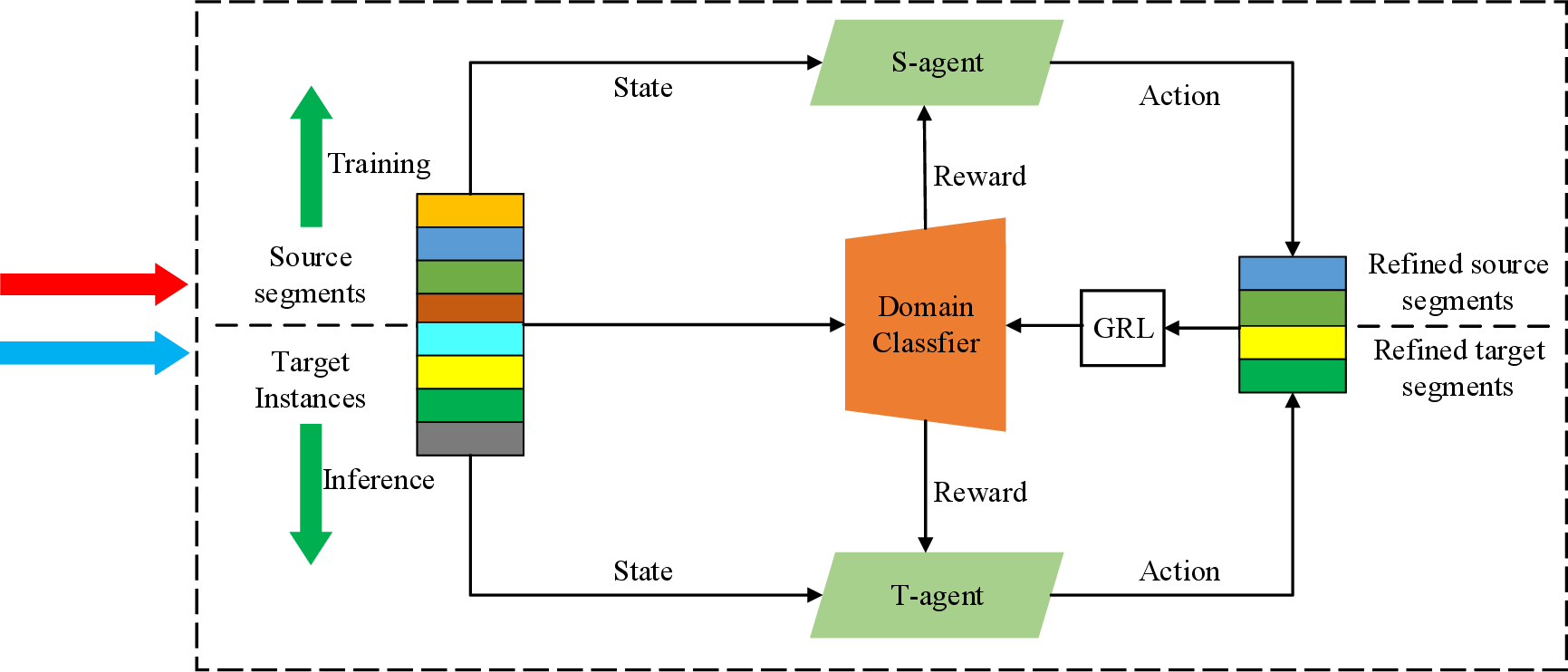}
        \vspace{5pt}
        \label{domain}
    \end{minipage}
    }
    \caption{Structure of proposed method MMIR: (a) An I3D~\cite{i3d} feature extractor in each modality is shared by both domains. The output feature vectors of I3D network are fed to the instance refinement and domain adversarial learning modules as well as action classifiers. (b) S-agent and T-agent are built for source and target domain, respectively. They take input feature vectors as state and make selections in the training instances to select out noisy samples. A domain classifier with GRL is optimized with refined instances, which gives rewards to agents according to their selections.}
    \label{model structure}
    \vspace{-15pt}
\end{figure}

The following sections give detailed explanations of our proposed method \emph{Multi-model Instance Refinement (MMIR)}. The architecture of our MMIR is shown in \figref{overall}, which is composed of a two-stream 3D convolutional feature extractor in two modalities: \textbf{RGB} and \textbf{Optical Flow} together with a domain discriminator and instance refinement module in each modality followed by an action classifier. The structure of domain discriminator and instance refinement module is depicted in \figref{domain}.

\subsection{Two-stream Action Recognition}

For multiple modalities, a feature fusion layer is applied after the action classifiers for summing the prediction scores from different modalities as shown in \figref{overall}. For input $\mathcal{X}$ with multiple modalities, we have $\mathcal{X}=(\mathcal{X}^1,\mathcal{X}^2,\ldots,\mathcal{X}^K)$, where $\mathcal{X}^k$ represents the input from the $k$-th modality. Therefore, we can define the classification loss as follows:
\begin{equation}
    \mathcal{L_{cls}} = \sum_{x\in S}-y\cdot \log\left (Softmax\left(\sum^K_{k=1}C^k\left(F^k\left(x^k\right)\right)\right)\right)
\end{equation}
where $y$ represents the class label, $C^k$ is the action classifier of the $k$-th modality, $F^k$ denotes the feature extractor of the $k$-th modality and $x^k$ represents source action segments from the $k$-th modality which are labeled. 

\subsection{Instance Refinement}

\begin{figure}[h]
\vspace{-10pt}
    \centering
    \includegraphics[width=3.5in]{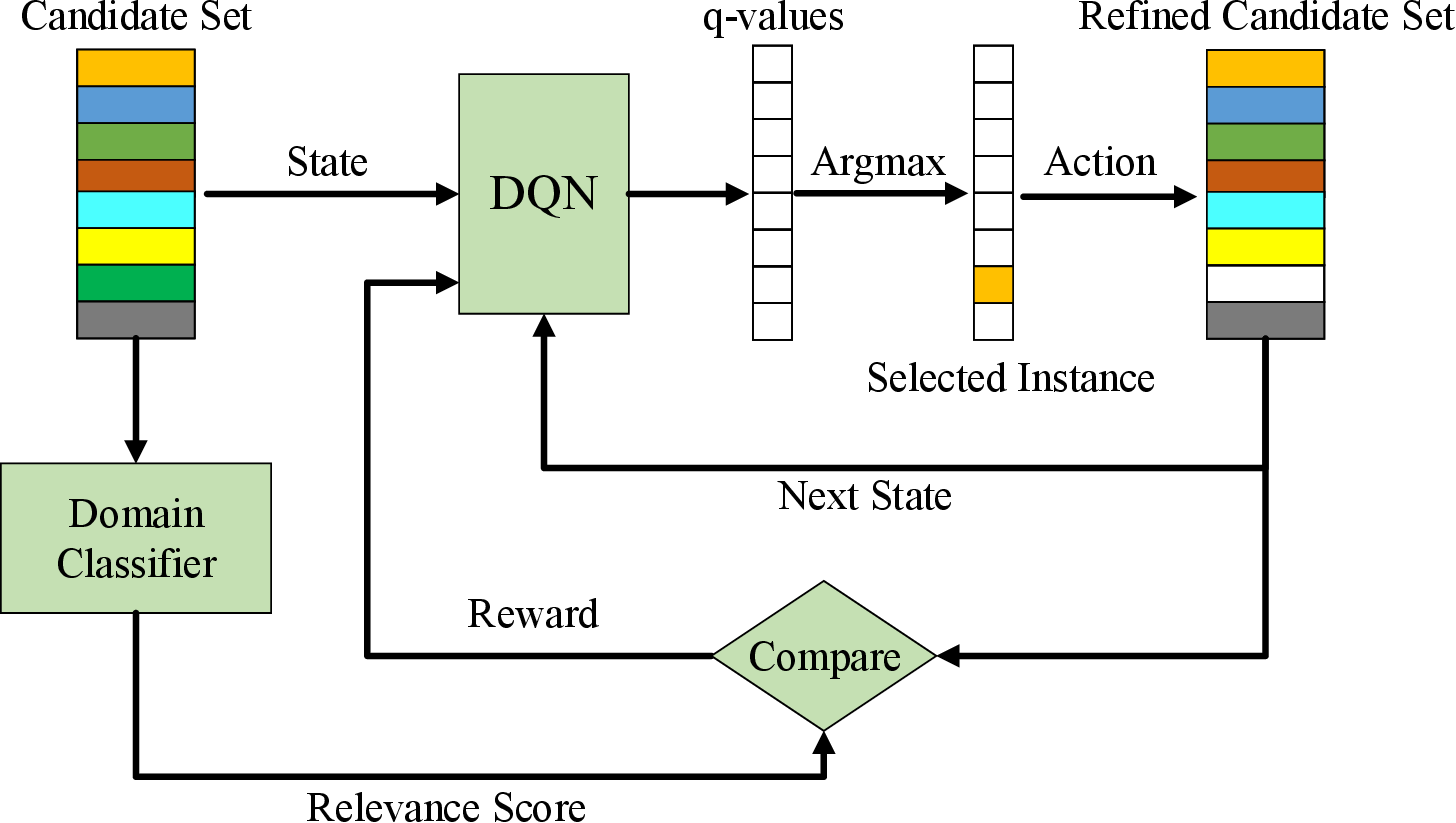}
    \vspace{-10pt}
    \caption{Workflow of the instance refinement module.}
    \label{dqn}
    \vspace{-15pt}
\end{figure}

 We visualize the overall workflow of instance refinement module in \figref{dqn}.
 In each modality, we select negative instances from the $\mathcal{i}$-th batch of action segments $\mathcal{F}^s_i$ and $\mathcal{F}^t_i$ in source and target domain, respectively. We divide a batch into several sub batches, namely, candidate set $\mathcal{F}_C$ for iterating the agents over more episodes. Therefore, we can have a total number of $\mathcal{E}$ candidate sets in a batch. Each episode is responsible for selecting out $E$ negative samples, thus the terminal time of an episode is defined as $E$. Take time $e$ in the selection process as an example, S-agent takes an action $\mathcal{A}^s_e$ by observing current state $\mathcal{S}^s_e$. Then, the current state is updated as $\mathcal{S}^s_{e+1}$ since an action segment has been selected out. In the meantime, S-agent would receive a reward $\mathcal{R}^s_e$ for taking action $\mathcal{A}^s_e$. After arriving at terminal time $E$, S-agent has done selection for this episode and the candidate set $\mathcal{F}_C$ is optimized as $\hat{\mathcal{F}_C}$. Then, the batch $\mathcal{F}^s_i$ would become $\hat{\mathcal{F}^s_i}$ at the end of the last episode, and similarly, we can reach a $\hat{\mathcal{F}^t_i}$ for $\mathcal{F}^t_i$ . 
 We give detailed illustrations on \emph{State}, \emph{Action}, \emph{Reward} and \emph{DQN Loss} in the following part of this section. 
\begin{itemize}
\vspace{-5pt}
\item
\textbf{State.}
Agents make selections on the level of candidate set and take feature vectors of all the action segments inside the candidate set as state. In this case, the state $\mathcal{S}^s_k$ of S-agent in the $\mathcal{k}$-th candidate set $\mathcal{C}^s_k$ could be defined as $\mathcal{S}^s_k =  [\mathcal{f}^s_{k,1},\mathcal{f}^s_{k,2},\mathcal{f}^s_{k,3},\ldots,\mathcal{f}^s_{k,\mathcal{N}_c}]\in\mathbb{R}^{d\times\mathcal{N}_c}$ where $\mathcal{f}^s_{k,n}$ is the feature vector of  $\mathcal{F}^s_{i,k.n}$ that has $d$ dimensions and $\mathcal{N}_c$ is the number of action segments inside a candidate set. Once an action segment $\mathcal{f}^s_{k,n}$ is selected out from $\mathcal{S}^s_k$ , it will be replaced by a $d$-dimensional zero vector to keep the state shape unchanged. This is the same for T-agent where $\mathcal{S}^t_k =  [\mathcal{f}^t_{k,1},\mathcal{f}^t_{k,2},\mathcal{f}^t_{k,3},\ldots,\mathcal{f}^t_{k,\mathcal{N}_c}]\in\mathbb{R}^{d\times\mathcal{N}_c}$ is the state and $\mathcal{f}^t_{k,n}$ is the feature vector of target action segment. 

\item
\textbf{Action.}
For a candidate set of size $\mathcal{N}_c$, we can have $\mathcal{N}_c$ actions to perform in each episode. Therefore, we can define the set of actions that can be taken by S-agent as $\mathcal{A}^s = \{1,2,\ldots,\mathcal{N}_c\}$ and T-agent as $\mathcal{A}^t = \{1,2,\ldots,\mathcal{N}_c\}$, which represents the index of the action segment that is to be selected out. The aim of the DQN agents is to maximize the accumulated reward of the actions taken. We define the accumulated reward at time $e$ as $\mathcal{R}_e=\sum_{t=e}^E\gamma^{t-e}r_e$, where $\gamma$ is the discount factor and $r_e$ represents the instant reward at time $e$. In DQN, we define a state-action value function to approximate the accumulated reward as $Q(\mathcal{S}_e,\mathcal{a}_e)$, where $\mathcal{S}_e$ denotes the state and $\mathcal{a}_e$ denotes the action taken at time $e$. For both modalities, $\mathcal{S}_e\in\{\mathcal{S}_e^s,\mathcal{S}_e^t\}$ and $\mathcal{a}_e\in\{\mathcal{A}^s,\mathcal{A}^t\}$. As shown in \figref{dqn}, DQN outputs a set of q-values corresponding to each action and chooses the optimal action which has the maximum q-value to maximize accumulated reward. This policy can be defined as follows: 
\begin{equation}
\hat{\mathcal{a}_e}=\max_{\mathcal{a}_e}Q(\mathcal{S}_e,\mathcal{a}_e).\label{qfunction}
\end{equation}

\item
\textbf{Reward.} Rewards given to agents are based on actions taken and the relevance of selected action segments to the opposite domain. To measure the relevance, we use the prediction results from domain classifier $D$. The domain logits of an action segment are processed by a sigmoid function and the relevance measure $\Delta(\mathcal{f})$  is defined as:
\begin{equation}
    \Delta(\mathcal{f})=
    \begin{cases}
Sigmoid(D(\mathcal{f})), & \mathcal{f}\in\hat{\mathcal{F^s}}\\
1-Sigmoid(D(\mathcal{f})), & \mathcal{f}\in\hat{\mathcal{F^t}}.
\label{rele}
\end{cases} 
\end{equation}
In Eq.\ref{rele}, we unify the relevance measure in both source and target domains by defining the domain label of source to be 0 and target to be 1. Then, the predefined threshold $\tau$ and the relevance measure $\Delta(\mathcal{f})$ can be compared to give rewards to agents according to the criterion defined below:
\begin{equation}
    r = 
    \begin{cases}
        1, & \Delta(\mathcal{f})<\tau, \\
        -1, & otherwise.
    \end{cases}
\end{equation}
This criterion is quite intuitive for an agent to recognize if it has made the right selection. Besides, we can set different thresholds for agents in different domains as $\tau^s$ and $\tau^t$ in source and target, respectively.

\item
\textbf{DQN Loss.}
For a DQN, the target output is defined as:
\begin{equation}
    y_e = r_e+\gamma\cdot\max_{\mathcal{a}_{e+1}}Q(\mathcal{S}_{e+1},\mathcal{a}_{e+1}|\mathcal{S}_e,\mathcal{a}_e)\label{dqnt}
\end{equation}
where $y_e$ represents the temporal difference target value of the Q function $Q(\mathcal{S}_e,\mathcal{a}_e)$. Based on this, the loss of DQN can be defined as:
\begin{equation}
    \mathcal{L_{q}}=\mathbb{E}_{\mathcal{S}_e,\mathcal{a}_e}[(y_e-Q(\mathcal{S}_e,\mathcal{a}_e))^2].
\end{equation}
Then, we can have the overall deep Q-learning loss defined as follows:
\begin{equation}
    \mathcal{L_{dqn}}=\sum^K_{k=1}(\mathcal{L_q^s}+\mathcal{L_q^t})_k    
\end{equation}
which is the sum of losses from S-agents and T-agents from all modalities.
\end{itemize}

\subsection{Domain Adversarial Alignment}

We realize feature alignment across domains in an adversarial way by connecting the domain discriminator with a GRL as shown in \figref{domain}. We apply a domain discriminator in each modality rather than using a single discriminator for all modalities after late fusion since aligning domains in a combined way might lead the network to focus on a less robust modality and lose the ability to generalize to other modalities~\cite{mmsada}. Then, we can define our domain adversarial loss as:
\begin{equation}
    \mathcal{L_{adv}}=\sum_{x^k\in\{S,T\}}-d\cdot\log\left(D^k\left(F^k\left(x^k\right)\right)\right)-\left(1-d\right)\cdot\log\left(1-D^k\left(F^k\left(x^k\right)\right)\right)
\end{equation}
where $d$ is the domain label, $D^k$ is the domain discriminator for the $k$-th modality, $F^k$ is the feature extractor of the $k$-th modality, and $x^k$ denotes the action segments from source domain or target domain of the $k$-th modality.

\subsection{Training}
With the losses defined in previous sections, we can have an overall loss function:
\begin{equation}
\label{tloss}
    \mathcal{L}=\mathcal{L_{cls}}+\mathcal{L_{dqn}}+\mathcal{L_{adv}}.
\end{equation}
For DQN agents, we use experience replay~\cite{ereplay} and $\epsilon$-greedy strategy~\cite{egreedy} during training. An experience replay pool to store actions, states, rewards, etc. is established for every agent, which ensures that data given to them is uncorrelated.
The $\epsilon$-greedy strategy introduces a probability threshold of random action $\epsilon$ to control whether an action is predicted by DQN agent or just randomly selected. This helps to balance the exploitation and exploration of an agent. The strategy is implemented as follows:
\begin{equation}
\hat{a_e}=
\begin{cases}
\max_{\mathcal{a}_e}Q(\mathcal{S}_e,\mathcal{a}_e) & \text{if} \ \lambda \ge \epsilon ,\\
a_e^{*} & otherwise,
\label{eps}
\end{cases} 
\end{equation}
where $\lambda$ is a random variable. If $\lambda$ is larger than $\epsilon$, the action would be predicted by the agents, or the action would be randomly chosen from the pool of actions.

\section{Experiments}

\subsection{Implementation Details}
\label{sec4.1}
\subsubsection{Dataset.}
 We use EPIC-Kitchens~\cite{epic} to set up the cross domain environment for fine-grained action recognition as it includes action segments captured in 32 different scenes. Following the domain split in~\cite{mmsada}, we sample videos taken in 3 different kitchens to form 3 different domains, which are denoted as D1, D2 and D3. 
 We have a total of 8 classes of action and according to~\cite{mmsada}, the distribution of training and testing samples from the 8 action classes are highly unbalanced. However, we use this unbalanced dataset to prove that our method could achieve competitive performance even on an unbalanced dataset since the unbalanced distribution of data makes domain adaptation more challenging.
\subsubsection{Model Architecture.}
\vspace{-10pt}
We set a two-stream I3D~\cite{i3d} feature extractor as our backbone. A trianing instance is composed of a temporal window of 16 frames sampled from an action segment. The domain discriminator $D^k$ in each modality takes in the feature vectors and flattens them to pass through a GRL and two fully connected layers with the dimension of a hidden layer to be 128 and an intermediate LeakyReLU activation function. For data augmentation, we follow the setup in~\cite{code} where random cropping, scale jittering and horizontal flipping are applied to training data. For testing data, only center cropping is applied.

\subsubsection{Hyperparameter and Training Setting.}
\vspace{-10pt}
The overall dropout rate of $F^k$ is set to 0.5 and a weight decay of $10^{-7}$ is applied for model parameters. We divide training process into two stages. In stage 1, our network is trained without domain discriminator and DQN agents. Then, the loss is optimized as follows:
    \begin{equation}
        \mathcal{L_{stage1}}=\mathcal{L_{cls}}.
    \end{equation}
    The learning rate of this stage is set to $0.01$ and the network is trained for 4000 steps. In the second stage, the domain discriminator and DQN agents are incorporated and the objective function for this stage is defined as:
    \begin{equation}
        \mathcal{L_{stage2}}=\mathcal{L_{cls}}+\mathcal{L_{adv}}+\mathcal{L_{dqn}}.
    \end{equation}
    The learning rate in this stage is reduced to $0.001$ and the model is further trained for 8000 steps. Note that for both stages, the action classifier is optimized only using labeled source data. 
    For the hyperparameters of DQN, we set the discount factor $\gamma=0.9$, $\epsilon$-greedy factor $\epsilon=0.5$, relevance threshold $\tau^s=\tau^t=0.5$, terminal time $E=1$ and candidate size $\mathcal{N}_c=5$. Besides, Adam~\cite{adm} optimizer is used for both stages and the batch size is set to 96 in stage 1 and 80 in stage 2, which is equally divided for source and target domains. It takes 6 hours to train our model using 4 NVIDIA RTX 3090 GPUs.


\subsection{Results}
\label{sec4.2}
For all the experimental results, we follow~\cite{mmsada} to report the average top-1 accuracy on target domain over the last 9 epochs during training. In the meantime, the experimental results of our model trained with only source data are reported as a lower limit. Also, we report results of supervised learning on target domain as the upper bound.

\begin{table}[h]
\vspace{-10pt}
    \centering
    \caption{Top-1 Accuracy of the experimental results of different baselines and our MMIR under different domain settings.}
    
    \belowrulesep=0pt
    \aboverulesep=0pt
    \resizebox{1\textwidth}{!}{
    \begin{tabular}{l|cccccccc}

   \toprule
   Method & D2 $\rightarrow$ D1 &D3 $\rightarrow$ D1 & D1$\rightarrow$ D2 & D3 $\rightarrow$ D2 & D1 $\rightarrow$ D3 & D2 $\rightarrow$ D3 & Mean \\
   \midrule
   Source-only & 42.5 & 44.3 & 42.0 & 56.3 & 41.2 & 46.5 & 45.5 \\
   MMD~\cite{udalong2} & 43.1 & 48.3 & 46.6 & 55.2 & 39.2 & 48.5 & 46.8 \\
   AdaBN~\cite{adbn} & 44.6  & 47.8 & 47.0 & 54.7 & 40.3 & 48.8 & 47.2\\
   MCD~\cite{mcd} & 42.1  & 47.9 & 46.5 & 52.7 & 43.5 & 51.0 & 47.3\\
  MM-SADA~\cite{mmsada} & 48.2  & 50.9 & 49.5 & 56.1 & 44.1 & 52.7 & 50.3 \\
  Kim \emph{et al.}~\cite{constra} & \textbf{49.5} & 51.5 & \textbf{50.3} & 56.3 & \textbf{46.3} & 52.0 & 51.0 \\
   \cmidrule{1-8}
   MMIR & 46.1  & \textbf{53.5} & 49.7 & \textbf{61.5} & 44.5 & \textbf{52.6} & \textbf{51.3}\\
   \cmidrule{1-8}
   Supervised & 71.7 & 74.0 & 62.8 & 74.0 & 62.8 & 71.7 & 69.5 \\
   
   \bottomrule
    \end{tabular}
    }

    \label{table:da}
    \vspace{-10pt}
\end{table}
We have a total of 6 domain adaptation settings based on D1, D2 and D3. We make a comparison between our method, several baseline methods and two state-of-the-art methods under every domain setting in Table \ref{table:da}. On average, our method outperforms all other state-of-the-art methods. MMIR has an overall performance improvement of $5.8\%$ compared with Source-only, $4.5\%$ compared with MMD~\cite{udalong2}, $4.1\%$ compared with AdaBN, $4.0\%$ compared with MCD~\cite{mcd}, $1.0\%$ compared with MM-SADA~\cite{mmsada} and $0.3\%$ compared with Kim \emph{et al.}~\cite{constra}.


\subsection{Ablation Study}
\label{sec.4.3}

\subsubsection{Effects of RL Agents.}
We evaluate the performance of reinforcement learning agents according to modality and domain. In particular, we investigate the effect of S-agent and T-agent only in the modality of RGB as this modality contributes more during the feature alignment process. The results are shown in Table \ref{table:agent} and we denote ``without'' as ``w/o''. We further elaborate on the results in the following part.
\begin{itemize}

    \item \emph{RGB vs Optical flow.} Our method without agents in RGB has a performance drop of $1.8\%$ while the case without agents in Optical flow has only a drop of $0.7\%$. This indicates that agents in RGB play a major part in refining feature alignment compared with that of Optical flow since RGB frames contain more spatial information while flow frames contain more temporal information which contributes less to the feature alignment process. 
    \item \emph{S-agent vs T-agent in RGB.} In RGB, when S-agent is removed, we can observe a performance drop of $0.3\%$. While by removing the T-agent, the performance drop is $0.8\%$. This shows that in the modality of RGB, T-agent weighs more than S-agent in alleviating the issue of negative transfer.
    \vspace{-20pt}
\end{itemize}

\begin{table}[H]
    \centering
    \caption{Ablation study of the effect of reinforcement learning agents on D2 $\rightarrow$ D1.}
    
    \belowrulesep=0pt
    \aboverulesep=0pt
    \resizebox{0.5\textwidth}{!}{
    \begin{tabular}{l|c}

   \toprule
   Method & D2 $\rightarrow$ D1 \\
   \midrule
   Source-only & 42.5  \\
   \cmidrule{1-2}
   MMIR (w/o) RGB agents & 44.3  \\
   MMIR (w/o) Flow agents & 45.4  \\
   MMIR (w/o) S-agent (RGB) & 45.8 \\
   MMIR (w/o) T-agent (RGB) & 45.3 \\
   \cmidrule{1-2}
   MMIR & \textbf{46.1} \\
   
   \bottomrule
    \end{tabular}
    }

    \label{table:agent}
    \vspace{-15pt}
\end{table}

\subsubsection{Overall Evaluation.}
In addition, we also evaluate the overall effect of our instance refinement strategy (\textbf{IR}) by comparing it with the case of \textbf{Adversarial-only} in Table \ref{table:ab}. We give a detailed illustration of our results as follows.
\begin{table}
\vspace{-15pt}
    \centering
    \caption{Overall evaluation of our instance refinement strategy.}
    \vspace{-5pt}
    \belowrulesep=0pt
    \aboverulesep=0pt
    \resizebox{1\textwidth}{!}{
    \begin{tabular}{lcc|cccccccc}

   \toprule
   Method & Adversarial & IR & D2 $\rightarrow$ D1 &D3 $\rightarrow$ D1 & D1$\rightarrow$ D2 & D3 $\rightarrow$ D2 & D1 $\rightarrow$ D3 & D2 $\rightarrow$ D3 & Mean \\
   \midrule
   Source-only & & & 42.5 & 44.3 & 42.0 & 56.3 & 41.2 & 46.5 & 45.5 \\
   \cmidrule{1-10}
   MMIR & \Checkmark & & 43 & 51.8 & 49.3 & 59.9 & 43.5 & 51.6 & 49.9 \\
   MMIR & \Checkmark & \Checkmark & \textbf{46.1}  & \textbf{53.5} & \textbf{49.7} & \textbf{61.5} & \textbf{44.5} & \textbf{52.6} & \textbf{51.3}\\
   
   \bottomrule
    \end{tabular}
    }

    \label{table:ab}
    
\end{table}

\begin{itemize}
\vspace{-20pt}
    \item \emph{Adversarial-only.} Compared with Source-only, an improvement of $4.4\%$ in top-1 accuracy can be observed. This shows that our domain adversarial alignment is effective in improving model performance on target domain through directly training a domain discriminator in an adversarial way.
    \item  \emph{Adversarial + IR.} Compared with Adversarial-only, it has an overall performance boost of $1.4\%$ and shows an improvement in every domain setting as depicted in Table \ref{table:ab}. This is an effective demonstration of the successful implementation of our instance refinement strategy and its capability to help alleviate negative transfer during cross-domain action recognition.
\vspace{-8pt}
\end{itemize}

\section{Conclusion}
We design a multi-modal instance refinement framework to help alleviate the problem of negative transfer during cross-domain action recognition. The reinforcement learning agents are trained to learn policies to select out negative training samples, thus resulting in a better-aligned feature distribution via domain adversarial learning. Experiments show that our method successfully addresses the negative transfer in multi-modal cross-domain action recognition and outperforms several competitive methods on a benchmark dataset. We believe in the future, it is worth conducting experiments on a spectrum of datasets to validate if our MMIR could be generalized to all use cases and even in different modalities such as text, speech, depth and so on.

\vspace{-12pt}
\subsubsection{Acknowledgements.} This work is supported by the Major Project for New Generation of AI under Grant No. 2018AAA0100400, National Natural Science Foundation of China No. 82121003, and Shenzhen Research Program No. JSGG20210802153537009.
\vspace{-8pt}
%
%
%
%
\bibliographystyle{splncs04}
\bibliography{paper}

\end{document}